\documentclass[letterpaper, 10 pt, conference]{ieeeconf}  % Comment this line out if you need a4paper

\IEEEoverridecommandlockouts                              % This command is only needed if 
\usepackage{graphics} % for pdf, bitmapped graphics files
\usepackage{graphicx}
\usepackage{algpseudocode,algorithm}

\usepackage{amsmath,amsfonts,amssymb}
\usepackage{color}
\usepackage{wrapfig}
\usepackage{subfigure}
\usepackage{multirow}
\usepackage{array}
\usepackage[table]{xcolor}
\usepackage{soul}
\usepackage{fixmath}
\newcolumntype{P}[1]{>{\centering\arraybackslash}p{#1}}
\newcolumntype{M}[1]{>{\centering\arraybackslash}m{#1}}

\DeclareMathOperator*{\argmin}{arg\,min}
\newcommand{\etal}{\textit{et al}.}

\newcommand{\quotes}[1]{``#1''}

\title{\LARGE \bf
Semi-Dense Visual Odometry for RGB-D Cameras \\ Using Approximate Nearest Neighbour Fields}

\author{Yi Zhou, Laurent Kneip and Hongdong Li\vspace{-3mm}% <-this % stops a space
\thanks{All the authors are with Research School of Engineering, the Australian National University. \{yi.zhou, laurent.kneip, hongdong.li\}@anu.edu.au. The research leading to these results is supported by Australian Centre for Robotic Vision. The work is furthermore
supported by ARC grants DE150101365.
Yi Zhou acknowledges the financial support from the China Scholarship
Council for his PhD Scholarship No.201406020098.}
}

\begin{document}

\maketitle
\thispagestyle{empty}
\pagestyle{empty}

%%%%%%%%%%%%%%%%%%%%%%%%%%%%%%%%%%%%%%%%%%%%%%%%%%%%%%%%%%%%%%%%%%%%%%%%%%%%%%%%
\begin{abstract}
This paper presents a robust and efficient semi-dense visual odometry solution for RGB-D cameras. The core of our method is a 2D-3D ICP pipeline which estimates the pose of the sensor by registering the projection of a 3D semi-dense map of the reference frame with the 2D semi-dense region extracted in the current frame. The processing is speeded up by efficiently implemented approximate nearest neighbour fields under the Euclidean distance criterion, which permits the use of compact Gauss-Newton updates in the optimization. The registration is formulated as a maximum a posterior problem to deal with outliers and sensor noises, and consequently the equivalent weighted least squares problem is solved by iteratively reweighted least squares method. A variety of robust weight functions are tested and the optimum is determined based on the characteristics of the sensor model. Extensive evaluation on publicly available RGB-D datasets shows that the proposed method predominantly outperforms existing state-of-the-art methods.
%A probabilistic formulation achieves robust estimation despite sensor noise and outliers. The resulting maximum a posteriori problem is replaced by an equivalent weighted least squares problem. A variety of robust weight functions are tested and the optimum is determined based on the characteristics of the sensor model. The processing is speeded up by efficiently implemented approximate nearest neighbour fields under the Euclidean distance criterion. This permits the use of compact Gauss Newton updates for tracking the pose of the camera. Extensive evaluation on publicly available RGB-D datasets shows that the proposed method predominantly outperforms the existing state-of-the-art.
\end{abstract}

%%%%%%%%%%%%%%%%%%%%%%%%%%%%%%%%%%%%%%%%%%%%%%%%%%%%%%%%%%%%%%%%%%%%%%%%%%%%%%%%
\section{Introduction}

Image-based estimation of camera motion, known as visual odometry (VO), plays a very important role in many robotic applications such as control and navigation of unmanned mobile robots, especially when no external navigation reference signal is available. Although a number of successful works have been presented over the past decade in this relatively mature field, the conclusion is that no method is all-powerful and working in any scenario. For example, salient feature based sparse methods such as~\cite{klein2007parallel,mur2015orb} do not work well when there is insufficient texture in the image for defining feature points. By taking advantage of all intensity information, direct methods like~\cite{tykkala2011direct,kerl2013robust,steinbrucker2011real,audras2011real} achieve better performance in textureless environments as long as the assumption of photometric consistency is sufficiently met. Engel \etal~\cite{engel2013semi,engel2014lsd} further improve the efficiency of~\cite{kerl2013robust} by using only photometric information around semi-dense region. Other systems such as~\cite{newcombe2011kinectfusion,whelan2012kintinuous,pomerleau2011tracking,pomerleau2013comparing} track the camera using an iterative closest point (ICP) algorithm over the depth information only. This, however, requires the presence of sufficient 3D structure and fails, for instance, in the situation of a planar scene. Furthermore, the ICP algorithm is always computationally expensive, and usually depends on GPU resources for real-time performance.

%In this work we present a semi-dense ICP based method for RGB-D cameras which take both robustness and computational efficiency into account. 
In this work we combine the merits of semi-dense processing and ICP based tracking. Compared to sparse methods, the proposed semi-dense method exploits the common structure of man-made environments, thus can handle relatively textureless situations. It ensures computational efficiency while being able to work in the degenerate case for ICP trackers (i.e. a single plane). Instead of applying a direct method which is sensitive to illumination changes, an accurate ICP inspired geometric framework is proposed. More precisely, the estimation of the camera pose at the current frame with respect to the reference frame is cast as a 2D-3D registration problem. The 3D part is given by a 3D semi-dense map defined in the reference frame, and the 2D part is given by the semi-dense region extracted in the current frame. Similar to the classical ICP framework which aims at aligning surfaces in 3D, our method aims at non-parametric curve-to-curve registration. To improve robustness against sensor noises and outliers, we apply an probabilistic model in the spirit of~\cite{kerl2013robust}. The resulting maximum a posteriori (MAP) problem is equivalent to a weighted least squares problem which can be solved using the iteratively re-weighted least squares (IRLS) method.
% By semi-dense regions, we mean pixels whose norm of gradient pass the threshold. Those pixels are reliable information because they are always stable, repeated and widely existed in images. Besides, the number of those pixels can always be controlled under 10,000 which is ideal deal for performing ICP on CPUs.
%

%%figure
\begin{figure}[t]
  \centering
  \includegraphics[width=0.70\columnwidth]{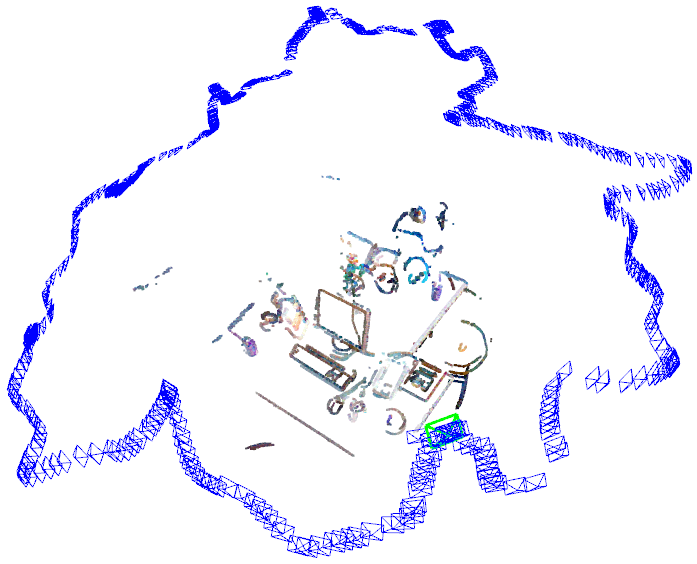}
  \caption{An illustration of the proposed semi-dense visual odometry. All reference frames are drawn in blue and the current frame in green. The colorful structures in the centre are the reconstruction of the semi-dense region observed by some of the reference frames.}
  \vspace*{-3.5 mm}
  \label{fig:pipeline}
\end{figure}

\vspace{1mm}
Our main contribution is four-fold:
\begin{itemize}
%\item A probabilistic formulation for 2D-3D semi-dense ICP based motion estimation.
%\item Careful choice of a proper robust weight function.
\item Introducing the idea of approximate nearest neighbour field, which permits the use of compact Gauss-Newton updates in the registration.
\item Exploring the optimal robust weight function for the probabilistically formulated, 2D-3D semi-dense ICP based motion estimation.
\item A real-time implementation running at 25 Hz on a laptop using only CPU resources.
\item Extensive evaluation under varying experimental conditions and varying algorithm setups (e.g., different methods for extracting the semi-dense region and reweighting residuals) and performance comparison against state-of-the-art solutions on publicly available datasets.
\end{itemize}

The paper is organized as follows. More related work is discussed in Section~\ref{Sec: Related work}. Section~\ref{Sec: Core concept} provides an in-depth review of geometric semi-dense 2D-3D registration. Then Section~\ref{Sec: Theory} presents the core of our new approach, including the idea of approximate nearest neighbour field and keys to robust motion estimation despite occlusion, noises and outliers. %The most robust weight function is identified by taking the characteristics of the sensor as well as the residual computation method into account. 
An overview of our framework is given in Section~\ref{Sec: Framework overview}, which is followed by extensive evaluation including the exploration of the best configuration and the comparison against state-of-the-art methods in Section~\ref{Sec: Evaluation}.
%%%%%%%%%%%%%%%%%%%%%%%%%%%%%%%%%%%%%%%%%%%%%%%%%%%%%%%%%%%%%%%%%%%%%%%%%%%%%%%%
\section{Related Work}
\label{Sec: Related work}

\textbf{Line, curve based and semi-dense methods:} Lines are alternative features to points and have been widely used in many VO and SLAM frameworks such as~\cite{eade2009edge,lu2015robustness}. One reason is that lines are abundant in man-made structures and environments, and do not depend on sufficient texture. Another reason is that line features are easily parametrized and included in a bundle adjustment (BA) pipeline~\cite{eade2009edge,klein2008improving} for the purpose of global optimization. However, straight lines are still not a general feature because object contours can be arbitrary curves in 3D space. Therefore, Nurutdinova \etal~presents a method which uses parametric curves as landmarks for motion estimation and BA~\cite{nurutdinova2015towards}. Futhermore, Engel \etal~apply direct method to semi-dense region~\cite{engel2013semi,engel2014lsd}, which fully utilizes the photometric information around all boundaries, edges and contours. The most relevant work to ours is~\cite{kuse2016robust}, which presents a direct edge alignment approach for 6-DOF tracking. They address the problem of non-differentiability of their Distance Transform (DT) based cost function by using a sub-gradient method. Conversely, we improve the differentiability of the cost function intrinsically and achieve more accurate results at a comparable computational cost.

\textbf{ICP:} The Iterative Closest Point (ICP) algorithm is a fundamental component of our method 
%of particular interest in the context of our method,
and it has been used exhaustively in 3D-3D registration problems. Typical issues when applying those methods are missing data, noise, outliers, and local minima in the registration process. Yang investigates globally optimal solutions to the point set registration problem~\cite{yang2013goicp}. However, this method is not efficient enough for real-time applications, where the frame-to-frame displacement remains small enough anyway for a successful application of local methods. The most related work is~\cite{BMVC2015_100} which applies ICP and distance transforms to semi-dense 3D-2D registration. Chebyshev/Chamfer distance field is chosen as an approximation of the Euclidean distance field to achieve real-time performance. Without discussing how to propagate the reference frame,~\cite{BMVC2015_100} stops at solving an absolute pose estimation problem rather than providing a full VO system.

\textbf{Photometric and hybrid registration methods:}
ICP algorithm and its close derivatives~\cite{pomerleau2011tracking,pomerleau2013comparing,newcombe2011kinectfusion,whelan2012kintinuous} still represent the methods of choice for real-time LIDAR tracking though sometimes expensive computational resources like GPU are necessary. The advent of RGB-D cameras has, however, led to a new generation of 2D-3D registration algorithms that exercise a hybrid use of both depth and RGB information. For instance, Steinbr{\"u}cker uses the depth information along with the optimized relative transformation to warp one RGB-D image to the next~\cite{steinbrucker2011real}, thus permitting direct and dense photometric error minimization. Similar idea is applied in~\cite{kerl2013robust,engel2013semi,engel2014lsd}.

\textbf{Robust M-estimators and IRLS:} When system noises and outliers are taken into account, M-estimators are popular choices for re-weighting the na\"{\i}ve least squares problem. The earliest tutorial about using different M-estimators in the application of conic fitting was given in~\cite{zhang1997parameter}. Recently, Aftab investigates the full range of robust M-estimators that are amenable to IRLS~\cite{aftab2015convergence}. In consideration of the great success of applying IRLS and M-estimators in motion estimation works such as~\cite{tykkala2011direct,kerl2013robust,engel2013semi}, we utilize it in our work as well.

%%%%%%%%%%%%%%%%%%%%%%%%%%%%%%%%%%%%%%%%%%%%%%%%%%%%%%%%%%%%%%%%%%%%%%%%%%%%%%%%
\section{Review of Geometric Semi-dense 2D-3D Registration}
\label{Sec: Core concept}

% Short explanation of the organization
This section reviews the basic idea behind geometric semi-dense 2D-3D registration. After a clear problem definition, we will review existing registration methods, and conclude with a brief summary of the open problems addressed in this paper.

%Finally, we conclude with details about the method and tools we use for solving the optimization problem.

%++++++++++++++++++++++++++++++++++++++++++++++++++++++++++
\subsection{Problem formulation}
\label{Subsec: Problem formulation}
% the goal

%%figure
\begin{figure}[b]
  \centering
  \vspace*{-3.5 mm}
  \subfigure[Image gradient's norm map.]{
  \includegraphics[width=0.45\columnwidth]{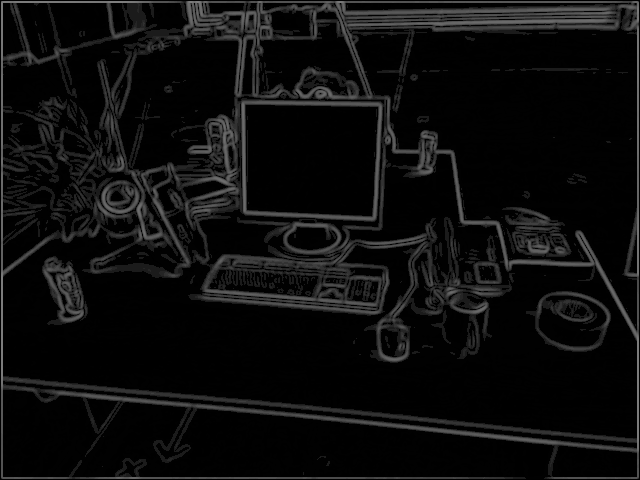}
  \label{fig: Image gradient map}}
  \subfigure[3D Semi-dense map.]{
  \includegraphics[width=0.45\columnwidth]{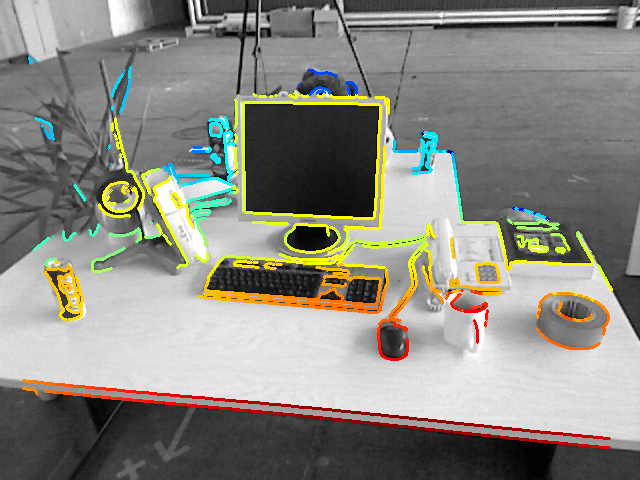}
  \label{fig: semi dense depth map}}
%   \subfigure[Reprojecting semi-dense depth map onto the current frame.]{
%   \includegraphics[width=0.28\columnwidth]{pic/align}
%   \label{fig: reprojection}}
  \caption{Image gradient is calculated in both horizontal and vertical direction at each pixel location. Euclidean norm of each gradient vector is calculated and illustrated in (a) (brighter means bigger while darker means smaller). Semi-dense region is obtained via thresholding the gradient's Euclidean norm map. By accessing the depth information of the semi-dense region, a 3D semi-dense map (b) is created,  in which hot colors refer to close while cold colors mean far. }
  \label{fig: problem definition}
\end{figure}

Let $\mathcal{P}^{\mathcal{F}}=\{\mathbf{p}_{i}^{\mathcal{F}}\}$ be a set of pixel locations in a frame $\mathcal{F}$ defining the so-called semi-dense region. As illustrated in Fig.~\ref{fig: problem definition}, it is obtained by thresholding the norm of the image gradient, which could, in the simplest case, originate from a convolution with Sobel kernels. Let us further assume that depth value $z_i$ for each pixel in the semi-dense region is available as well. In the preregistered case, they are simply obtained by looking up the corresponding pixel location in the associated depth image. For each pixel, a local patch ($5 \times 5$ pixels) is visited and the smallest depth is selected in the case of a depth discontinuity\footnote{The depths of all pixels in the patch are sorted and clustered based on a simply Gaussian noise assumption. If there exists a cluster center that is closer to the camera, the depth value of the current pixel will be replaced by the depth of that center. This circumvents resolution loss and elimination of fine depth texture.}. This operation ensures that we always retrieve the foreground pixel despite possible misalignments caused by calibration errors or asynchronous measurements under motion. An example result is indicated in Figure \ref{fig: semi dense depth map}. We furthermore assume that both the RGB and the depth camera are fully calibrated (intrinsically and extrinsically). Thus we have accurate knowledge about a world-to-camera transformation function $\pi(\lambda \mathbf{f}_{i})=\mathbf{p}_{i}$ projecting any point along the ray defined by unit vector $\mathbf{f}_i$ onto the image location $\mathbf{p}_i$. The inverse transformation $\pi^{-1}(\mathbf{p}_{i})=\mathbf{f}_{i}$ transforming points in the image plane into unit direction vectors located on the unit sphere around the center of the camera is also known. If the RGB image and the depth map are already registered, the extrinsic parameters can be omitted. Our discussion from now on will be based on this assumption.

% If RGB image and depth map are already registered, the depth information for the semi-dense regions can be directly obtained by accessing the corresponding pixel of the depth image 

% which semi-dense depth map (SDDM) is defined by obtaining 

% can be regarded as a collection of 3D curves defined in a reference frame.
% If 

% Let $\mathcal{F}_{ref}:=\{f_{ref}^{rgb},f_{ref}^{depth}\}$ be a reference frame in which both RGB and depth information are available and l

% The depth information for the semi-dense regions can be directly obtained by accessing the corresponding pixel of the depth image if rgb image and depth map are already registered.

% the goal of the motion estimation consists retrieving the pose of a neighbouring frame $\mathcal{F}_{k}:=\{f_{k}^{rgb}\}$ given by position $\mathbf{t}$ and orientation $\mathbf{R}$ such that the reprojection of the semi-dense depth map (SDDM) in reference frame $\mathcal{F}_{ref}$ aligns well with the semi-dense regions (2D) extracted in frame $\mathcal{F}_{k}$. 
% preliminary

% Describe the algorithm and discuss the Euclidean Vector Map
Consider the 3D semi-dense map (defined in the reference frame $\mathcal{F}_\textup{ref}$) as a curve in 3D, and its projection into the current frame $\mathcal{F}_{k}$ as a curve in 2D. The goal of the registration step consists of retrieving the pose at the current frame $\mathcal{F}_{k}$ (namely its position $\mathbf{t}$ and orientation $\mathbf{R}$) such that the projected 2D curve aligns well with the semi-dense region $\mathcal{P}^{\mathcal{F}_{k}}$ extracted in the current frame $\mathcal{F}_{k}$. Note that---due to perspective transformations---this is of course not a one-to-one correspondence problem. Also note that we parametrize our curves by a set of points originating from pixels in an image. %While there are alternatives to this, the optimization objectives outlined in this work will remain mostly valid.

%++++++++++++++++++++++++++++++++++++++++++++++++++++++++++
\subsection{ICP-based motion estimation}
\label{subsec: icp based motion estimation}

The problem can be formulated as follows. Let
\begin{equation}
\mathcal{S}^{{\mathcal{F}}_\textup{ref}}
= \bigg\{ \mathbf{s}_{i}^{\mathcal{F}_\textup{ref}} \bigg\}
= \bigg\{ d_{i}^{\mathcal{F}_\textup{ref}}\pi^{-1}\big(\mathbf{p}_{i}^{\mathcal{F}_\textup{ref}}\big) \bigg\}
\label{eq:sdmap}
\end{equation}
denote the 3D semi-dense map in reference frame $\mathcal{F}_\textup{ref}$, where $d_{i} = \frac{z_{i}}{\mathbf{f}_{i,3}}$ denotes the distance of point $\mathbf{s}_{i}$ to the optical center. Its projection into current frame $\mathcal{F}_{k}$ results in the semi-dense region
\begin{equation}
\mathcal{O}^{\mathcal{F}_{k}}
= \bigg\{ \mathbf{o}_{i}^{\mathcal{F}_{k}} \bigg\}
= \bigg\{ \pi \Big( \mathbf{R}^\textup{T} \big( \mathbf{s}_{i}^{\mathcal{F}_\textup{ref}} - \mathbf{t} \big) \Big) \bigg\}.
\end{equation}
We define
\begin{equation}
n(\mathbf{o}_{i}^{\mathcal{F}_{k}}) = \underset{\mathbf{p}_{j}^{\mathcal{F}_{k}} \in \mathcal{P}^{\mathcal{F}_{k}}}{\operatorname{argmin}} \| \mathbf{p}_{j}^{\mathcal{F}_{k}} - \mathbf{o}_{i}^{\mathcal{F}_{k}} \|
\label{eq:nn}
\vspace{-2mm}
\end{equation}
to be a function that returns the nearest neighbour of $\mathbf{o}_{i}^{\mathcal{F}_{k}}$ in $\mathcal{P}^{\mathcal{F}_{k}}$ under the Euclidean distance metric. The overall objective of the registration is to find

\begin{equation}
\hat{\mathbold{\theta}}=\underset{\mathbold{\theta}}{\operatorname{argmin}} \sum_{i=1}^{N} \|  \mathbf{o}_{i}^{\mathcal{F}_{k}} - n(\mathbf{o}_{i}^{\mathcal{F}_{k}})\|^{2},
\label{eq:sol1}
\vspace{-1mm}
\end{equation}
where $\mathbold{\theta}:=[c_1,c_2,c_3,t_x,t_y,t_z]^\textup{T}$ represents the parameter vector that defines the pose of the camera. $c_1,c_2,c_3$ are Cayley parameters~\cite{cayleyparameter} for orientation $\mathbf{R}$\footnote{Note that the orientation is always optimized as a change with respect to the previous orientation in the reference frame. The chosen Cayley parametrization therefore is equivalent to the local tangential space at the location of the previous quaternion orientation and, therefore a viable parameter space for local optimization of the camera pose.}, and $\mathbf{t} = [t_x,t_y,t_z]^\textup{T}$. The above objective is of the same form as the classical ICP problem, which alternates between finding approximate nearest neighbours and then register those putative correspondences, except that in the present case, the correspondences are between 2D and 3D entities. A very similar objective function has been already exploited by \cite{BMVC2015_100} for robust semi-dense 2D-3D registration in a hypothesis-and-test scheme. It proceeds by iterative sparse sampling and closed-form registration of approximate nearest neighbours.

%++++++++++++++++++++++++++++++++++++++++++++++++++++++++++
\subsection{Distance fields}
\label{subsec:Distance fields}
As already outlined in \cite{BMVC2015_100}, the repetitive explicit search of nearest neighbours is too slow even in the case of robust sparse sampling. This is due to the fact that all distances need to be computed in order to rank the hypotheses, and this would again require an exhaustive nearest neighbour search. This is where distance transforms come into play. The explicit location of a nearest neighbour does not necessarily matter in order to evaluate the optimization objective (\ref{eq:sol1}), the distance alone may already be sufficient. Therefore, we can pre-process the semi-dense region in the current frame and derive an auxiliary image in which the value at every pixel simply denotes the Euclidean distance to the nearest point in the original semi-dense region. Euclidean distance fields can be computed very efficiently using region growing techniques. Chebychev distance is an alternative when faster performance is required. %If even faster alternatives are required, the Euclidean distance can be replaced by the Chebychev distance.
For further information, the interested reader is referred to~\cite{fabbri20082d}.

Let us define the function $d(\mathbf{o}_{i}^{\mathcal{F}_{k}})$ that retrieves the distance to the nearest neighbour by simply looking up the value at $\mathbf{o}_{i}^{\mathcal{F}_{k}}$ inside the chosen distance field. The optimization objective (\ref{eq:sol1}) can now easily be rewritten as\vspace{-0.5mm}
\begin{equation}
\hat{\mathbold{\theta}}=\underset{\mathbold{\theta}}{\operatorname{argmin}} \sum_{i=1}^{N}  d(\mathbf{o}_{i}^{\mathcal{F}_{k}})^{2}.
\vspace{-0.5mm}
\label{eq:sol2}
\end{equation}
Note that, in order to emulate a smooth optimization objective and bypass the effects of image discretization, the distances in the field are sampled using bilinear interpolation.

There are a few problems with objective (\ref{eq:sol2}):
\begin{itemize}
\item It is the sum of squared residual distances. The residual distance is a positive entity which means that it is hard to optimize by techniques other than gradient descent like methods\footnote{The values of the residual distance are always positive, which makes Gauss-Newton method not applicable. We discuss how to enable Gauss-Newton method by introducing a novel alternative to Euclidean distance field in the following section.}.
%the only valid way to minimize this objective is by applying the gradient descent technique.
Despite that it may have good convergence properties, it is known to be slow due to the cascaded update procedure, which may for instance involve a bisectioning line-search along the gradient direction.
\item As very well explained in \cite{nurutdinova2015towards}, the distance transform may easily lead to wrong registrations. For instance, if only a part of a model curve is observed in the current frame, the corresponding distance field may easily converge in a wrong location, even if only translational displacements in the image plane are taken into account. A detailed illustration of this problem is given in Fig. 3 of \cite{nurutdinova2015towards}. In their work, they solve the problem through a variable lifting strategy, which however blows up the space of optimized parameters quite significantly.
\item Even in the absence of the above two problems, a simple continuous minimization of the L2-norm of the residual distances would simply fail because it is easily affected by outlier associations. In \cite{BMVC2015_100}, this problem is circumvented by switching to the L1-norm of the residual distances. While a direct continuous minimization of the L1-norm is practically feasible, it remains conceptually wrong as claimed in~\cite{kuse2016robust} that the plain residual distance is not necessarily differentiable around zero.
\end{itemize}
The following section will address these problems one by one.

%%%%%%%%%%%%%%%%%%%%%%%%%%%%%%%%%%%%%%%%%%%%%%%%%%%%%%%%%%%%%%%%%%%%%%%%%%%%%%%%
\section{Theory}
\label{Sec: Theory}

The idea of approximate nearest neighbour field is first discussed, which enables registration through only few Gauss-Newton iterations. Then we introduce the gradient directions to project the residuals, thus leading to correct registration even though only part of a model is observed. %Afterwards, we demonstrate how to robustify the motion estimation by formulating the problem as a MAP estimation and translate it into a weighted least squares problem. 
Afterwards, we follow the probabilistic formulation given by~\cite{kerl2013robust} and solve its equivalent weighted least squares problem. Finally, the sensor model is learned to determine the optimal weight function. Note that our tracking approach has similarities with \cite{tarrio15}. However, our approximate nearest neighbour field obey the Euclidean distance metric, and we provide a more concise derivation of the Gauss-Newton update steps including robustification against outliers.

%++++++++++++++++++++++++++++++++++++++++++++++++++++++++++
\subsection{Approximate Nearest Neighbour Field}

\begin{figure}[t]
  \centering
  \vspace{2mm}
  \subfigure[ANNF.]{
  \includegraphics[width=0.35\columnwidth]{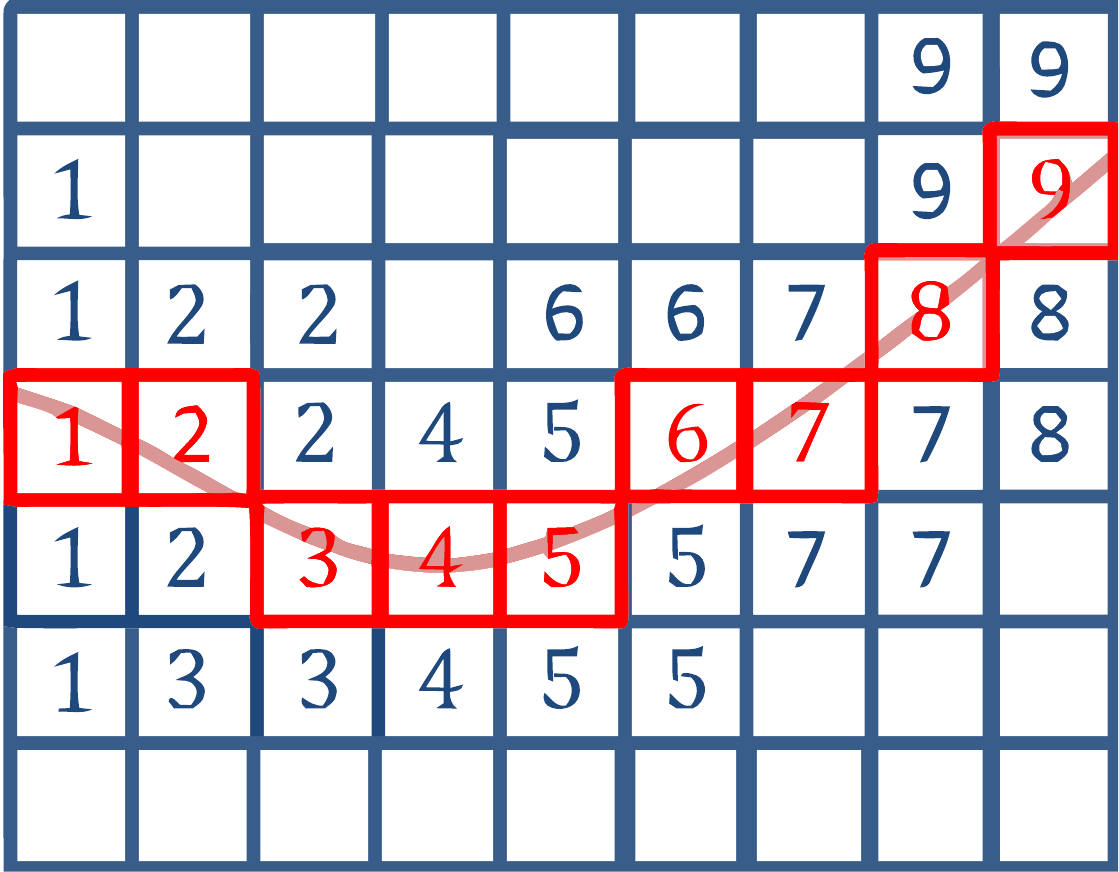}
  \label{fig: euclidean vector map}}
  \subfigure[Projected residual.]{
  \includegraphics[width=0.35\columnwidth]{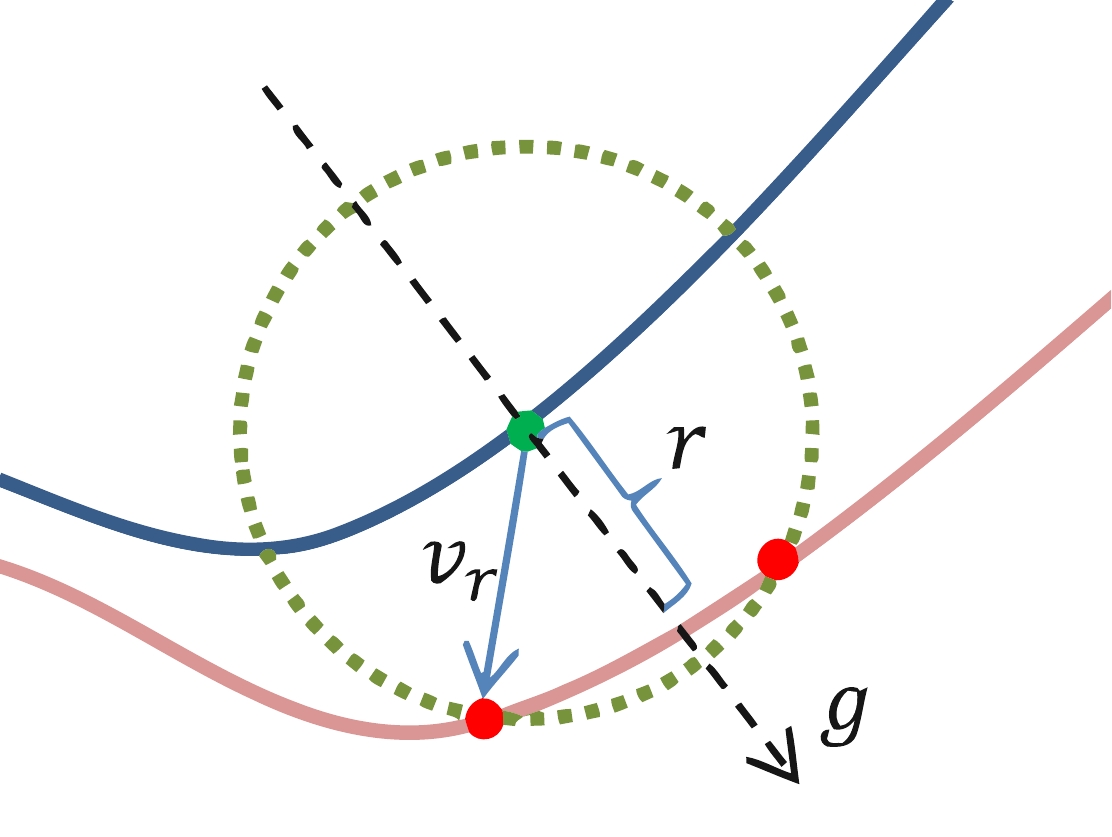}
  \label{fig: gradient distance}}
  \caption{Illustration of the Approximate Nearest Neighbour Field and the projected residual. The red numbers in (a) represent the index of the points on the red curve. Given the coordinate of a point in the map, its closest point can be easily accessed by checking the blue number, thus leading to the residual vector $\mathbf{v}_r$. The projected distance $r$ is finally calculated by projecting $\mathbf{v}_r$ onto the direction of the local gradient $g$.}
  \vspace{-3.5mm}
  \label{fig: curve registration}
\end{figure}

%While the introduction of a distance field enables us to efficiently retrieve the distance to the nearest point on a curve, it obscures the actual location of it. The latter would however be very useful in order to retrieve a full (signed) reprojection error, and thus unlock the possibility of performing efficent Gauss-Newton updates.
As discussed in \ref{subsec:Distance fields}, a full (signed) residual is needed to make the Gauss-Newton updates applicable. Thus, we replace Euclidean distance field with one that can retrieve the exact location of the nearest point on a curve. There is a straightforward alternative to the commonly used distance field that maintains all necessary information for computing full residuals, namely an \textit{Approximate Nearest Neighbour Field (ANNF)}. An ANNF is given by a $w\times h\times 2$ integer matrix, where $w$ denotes the width of the image, and $h$ its height. The integers at coordinates $(x,y,:)$ simply denote the pixel index of the nearest neighbour, rather than the distance to it. An example of an ANNF is illustrated in Fig.~\ref{fig: euclidean vector map}.

What is perhaps surprising is that the ANNF can be computed equally efficiently than the distance field. The reason for this is simply given by the functioning of efficient Euclidean distance field extraction algorithms. They perform region growing starting from the semi-dense region itself. The border of the growing region updates and propagates a reference to the closest point in the seed region (i.e. the original semi-dense region). Extracting a distance field or an ANNF is simply a matter of what piece of information is retained.

Using the ANNF, the function $n(\mathbf{o}_{i}^{\mathcal{F}_{k}})$ from (\ref{eq:nn}) now boils down to a trivial look-up. This enables us to again go back to objective (\ref{eq:sol1}), and attempt a solution via Gauss-Newton updates. Let us define the residuals
\begin{equation}
  \mathbf{v}=\left[ \begin{matrix}
      \mathbf{o}_1^{\mathcal{F}_k} - n(\mathbf{o}_1^{\mathcal{F}_k})\\
      \ldots \\
      \mathbf{o}_N^{\mathcal{F}_k} - n(\mathbf{o}_N^{\mathcal{F}_k})
  \end{matrix} \right]_{2N \times 1}.
\label{eq:reproj_error}
\end{equation}
By using (\ref{eq:reproj_error}) in (\ref{eq:sol1}), our optimization objective can be reformulated as
\begin{equation}
  \hat{\mathbold{\theta}} = \underset{\mathbold{\theta}}{\operatorname{argmin}} \| \mathbf{v} \|^{2}.
\label{eq:sol3}
\end{equation}

Supposing that $\mathbf{v}$ were a linear expression of $\mathbold{\theta}$, it is clear that solving (\ref{eq:sol3}) would be equivalent to solving $\mathbf{v}(\mathbold{\theta})=\mathbf{0}$. The idea of Gauss-Newton updates (or iterative least squares) consists of iteratively performing a first-order linearization of $\mathbf{v}$ about the current value of $\mathbold{\theta}$, and then each time improve the latter by solving the resulting linear least squares problem. The linear problem to solve in each iteration therefore is given by
\begin{equation}
  \mathbf{v}(\mathbold{\theta}_{i})+\left. \frac{\partial \mathbf{v}(\mathbold{\theta})}{\partial \mathbold{\theta}}\right|_{\mathbold{\theta}=\mathbold{\theta_i}} \mathbold{\Delta} = \mathbf{0},
  \label{eq:itlinsol}
  \vspace{-1mm}
\end{equation}
and, using $\mathbf{J}=\left. \frac{\partial \mathbf{v}(\mathbold{\theta})}{\partial \mathbold{\theta}}\right|_{\mathbold{\theta}=\mathbold{\theta_i}}$, its solution is given by
\begin{equation}
  \mathbold{\Delta} = - (\mathbf{J}^\textup{T}\mathbf{J})^{-1} \mathbf{J}^\textup{T}\mathbf{v}(\mathbold{\theta}_{i}).
\end{equation}
The motion vector is finally updated as $\mathbold{\theta}_{i+1} = \mathbold{\theta}_{i}+\mathbold{\Delta}$.

While this may sound straightforward, there is one element that requires particular attention. The nearest neighbour of each point should remain fixed in each round of iterative least squares. This statement particularly addresses the (numerical) Jacobian computation, as even tiny variations of $\mathbf{o}_{i}^{\mathcal{F}_k}$ can easily lead to a potentially substantial change of the nearest neighbour $n(\mathbf{o}_{i}^{\mathcal{F}_{k}})$ (e.g. from a point in one curve to another point in a completely different curve). We circumvent this problem by fixing the nearest neighbours during the Jacobian computation. The Jacobian $\mathbf{J}$ simply becomes
\begin{equation}
  \mathbf{J} = \left[ \begin{matrix}
    \left( \frac{\partial \mathbf{o}_1^{\mathcal{F}_k}}{\partial \mathbold{\theta}} \right)^\textup{T} &
    \ldots &
    \left( \frac{\partial \mathbf{o}_N^{\mathcal{F}_k}}{\partial \mathbold{\theta}} \right)^\textup{T}
  \end{matrix} \right]_{\mathbold{\theta}=\mathbold{\theta_i}}^\textup{T}.
  \label{eq:newjacobian}
  \vspace{-1mm}
\end{equation}

% Due to introducing the Euclidean vector map, the non-parametric curve-to-curve registration performs better than point-to-point registration which utilizes standard Euclidean distance field. An experiment is provided in the evaluation part show demonstrating the improvement.
% Instead of using a standard Euclidean distance field in which the closest distance for each point $\mathbf{o}_i^{\mathcal{F}_k}$ is recorded as a scalar, we develop a new strategy which is called Euclidean vector map. The closet point is still found in Euclidean metric, however a vector $\mathbf{v}_r$ that is originating from $\mathbf{o}_i^{\mathcal{F}_k}$ and pointing to its closest point $n(\mathbf{o}_i^{\mathcal{F}_k})$ is recorded. This residual vector enable us to project the residual onto the local gradient direction. The sum of those projected distance represents the orthogonal distance between the two under-aligned curves. In other wordsk

% In order to register the two curves shown in Fig~\ref{fig: curve registration}

%%%%figure: ANNF%%%%
%We present a new distance metric called Euclidean distance vector map illustrated in Fig~{fig:}. Compared to standard Euclidean distance field, the Euclidean vector map enables each points to keep the coordinate of its closest point (under Euclidean metric) as constant during the calculation of Jacobian. This new distance metric makes the energy function more smooth.

%++++++++++++++++++++++++++++++++++++++++++++++++++++++++++
%\vspace{-3mm}
\subsection{Projection of residual vectors}

While the fixation of the nearest neighbours during the Jacobian computation has clear benefits, it also leads to one further problem. Imagine a case where we have to register one long horizontal and one short vertical line in the image plane, and there are only two degrees of freedom. The horizontal line is already registered, but the vertical one not yet. A shift along the horizontal axis would solve the problem, however, the Jacobian will not provoke an overall error reduction along this dimension. This is because---with fixed nearest neighbours---the ``registered'' points along the horizontal edge may lead to spurious residual errors for any horizontal shift that ultimately outweighs the error reduction along the short vertical line.%growing residual errors for any horizontal shift.

As shown in Fig.~\ref{fig: gradient distance}, we solve this problem by projecting the residual vectors onto the local gradient directions. The new residual is given by
\begin{equation}
  \mathbf{r}=\left[ \begin{matrix}
      \left( \mathbf{o}_1^{\mathcal{F}_k} - n(\mathbf{o}_1^{\mathcal{F}_k}) \right)^\textup{T}\mathbf{g}(\mathbf{p}_1^{\mathcal{F}_{\text{ref}}})\\
      \ldots \\
      \left( \mathbf{o}_N^{\mathcal{F}_k} - n(\mathbf{o}_N^{\mathcal{F}_k}) \right)^\textup{T}\mathbf{g}(\mathbf{p}_N^{\mathcal{F}_{\text{ref}}})
  \end{matrix} \right]_{N \times 1},
\label{eq:reproj_error2}
\end{equation}
where the gradient of the registered point in the reference frame is denoted by $\mathbf{g}(\mathbf{p}_i^{\mathcal{F}_{\text{ref}}})$, and remains fixed throughout the optimization. This is only an approximation of the local curve gradient in the current frame, which is sufficiently valid under the assumption that the frame-to-frame transformation---and notably the rotation about the principal axis of the camera---is small enough. Also, while the residual errors have now become scalars again, they remain signed entities, and thus Gauss-Newton remains applicable. The new Jacobian is finally given by
\begin{equation}
  \mathbf{J} = \left[ \begin{matrix}
    \Big( \frac{\partial \big( \mathbf{g}(\mathbf{p}_1^{\mathcal{F}_{\text{ref}}} )^\textup{T}\mathbf{o}_1^{\mathcal{F}_k}\big)}{\partial \mathbold{\theta}} \Big)^\textup{T} &
    \ldots &
    \Big( \frac{\partial \big( \mathbf{g}(\mathbf{p}_N^{\mathcal{F}_{\text{ref}}} )^\textup{T}\mathbf{o}_N^{\mathcal{F}_k}\big)}{\partial \mathbold{\theta}} \Big)^\textup{T}
  \end{matrix} \right]_{\mathbold{\theta}=\mathbold{\theta_i}}^\textup{T}.
  \label{eq:newjacobian}
\end{equation}
Note that the projection of residual vectors onto the local gradient direction also helps to better approximate the orthogonal distance between curves, and thus address the problem raised in \cite{nurutdinova2015towards}---how to avoid wrong registrations in the case where some of the curves are observed partially.

%++++++++++++++++++++++++++++++++++++++++++++++++++++++++++
\subsection{Robust motion estimation}

% Objective (\ref{eq:sol3}) can be solved via Iterative Least Squares (ILS). However, it is well known that a na\"{\i}ve LS solver is susceptible to severe noise and outliers. Inspired by~\cite{kerl2013robust}, we reformulate the problem as a likelihood function of camera motion such that noise and outliers are properly dealt with.
%
%The residual of the i-th point $s_i^{\mathcal{F}_{ref}}$ is defined as
%the projection of the geometric distance onto the local gradient direction, i.e.,%between its reprojection $o_i^{\mathcal{F}_{k}}$ and the closest point $n( o_{i}^{\mathcal{F}_k} )$, i.e.,
%
%\begin{equation}
%r_i(\mathbold{\theta}):= \left[
%   \mathbf{o}_{i}^{\mathcal{F}_{k}} - 
%   n(\mathbf{o}_{i}^{\mathcal{F}_{k}})
%  \right]^{T}
%  \mathbf{g}(\mathbf{p}_i^{\mathcal{F}_{\text{ref}}} ),
%\end{equation}
%
% %
% \begin{equation}
% p(\mathbf{r} \vert \mathbf{R,t}) = \prod_{i} p(r_i \vert \mathbf{R,t})
% \label{eq:iid assumption}
% \end{equation} 
% %
%

From a probabilistic point of view, the motion would be estimated by maximizing the posteriori $p(\mathbold{\theta} \vert \mathbf{r})$ in the presence of noises. Following the derivation in~\cite{kerl2013robust}, the Maximum A Posteriori (MAP) problem is translated into the weighted least squares minimization problem,
\begin{equation}
\mathbold{\theta} = \argmin_{\mathbold{\theta}} \sum_i \omega(r_i)(r_i(\mathbold{\theta}))^{2}.
\label{eq: weighted least squares}
\end{equation}
The weight is defined as $\omega (r_i) = -\frac{1}{2r_i}\frac{\partial \log p(r_i \vert \mathbold{\theta})}{\partial r_i}$, which is a function of the sensor model $p(r_i \vert \mathbold{\theta})$. IRLS is used for solving (\ref{eq: weighted least squares}) and we discuss how to determine the optimal weight function $\omega(\cdot)$ by learning the statistical characteristics of the sensor model in the next section. 

\begin{figure}[t]
  \centering
  \vspace{2.5mm}
  \includegraphics[width=0.85\columnwidth]{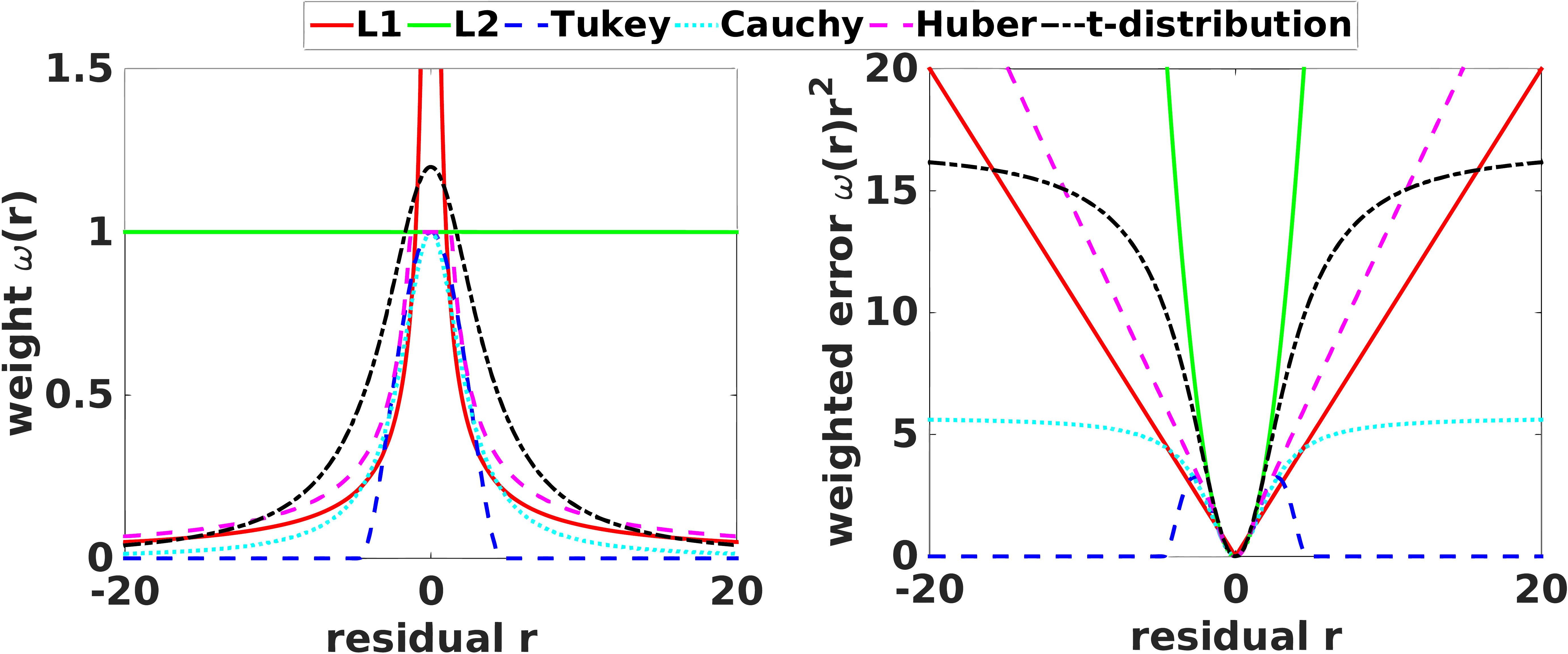}
  \caption{Illustration of several robust weight functions and corresponding weighted squared errors. The parameters used are from \cite{zhang1997parameter}.}
  \vspace{-5mm}
  \label{fig: Robust weight function}
\end{figure}

%%%%figure: sensor models%%%%
\begin{figure}[b]
  \centering
  \includegraphics[width=0.95\columnwidth]{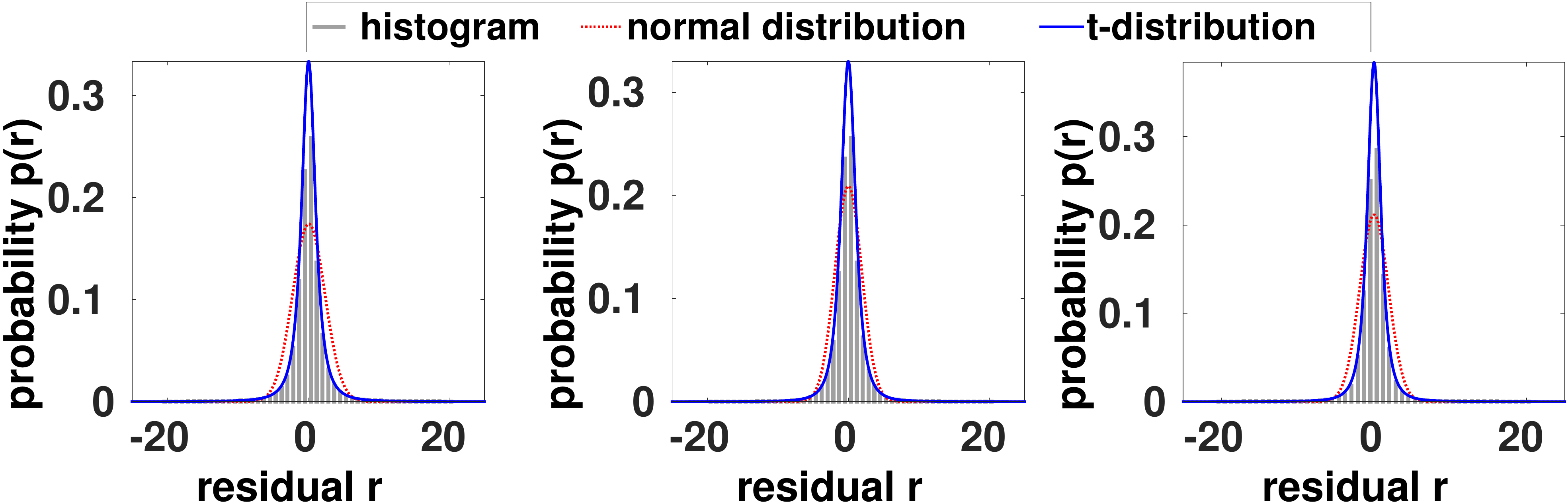}
  \caption{Sensor model $p(r)$ is obtained by fitting the histogram with certain probabilistic distribution. T-distribution gives best fitting result for all selected sequences fr1/floor, fr2/desk, fr3/structure\_texture\_near.}
  \label{fig: fitted sensor model}
\end{figure}

%+++++++++++++++++++++++++++++++++++++++++++++++
\subsection{Learning the sensor model}

%IRLS is a widely used technique for solving parameter estimation problems from noisy data such as rotation averaging, point cloud alignment and conic fitting, etc. 
% There exists several work like~\cite{zhang1997parameter,aftab2015convergence} discussing the properties of various robust cost function. 
We investigate several of the most widely used robust weight functions. %which has been discussed in~\cite{zhang1997parameter}. %which are summarized in Table~\ref{table: robust weight functions}. 
They and their corresponding weighted squared errors are illustrated in Fig.~\ref{fig: Robust weight function}. The interested reader can find more details in~\cite{zhang1997parameter,aftab2015convergence}.

% Seen from Fig.~\ref{fig: Robust weight function}, weighted squared error functions based on different $\omega(r)$ present different shapes which will lead to different converging behavior. 

The choice of the weight function depends on the statistics of the residual, which is identified in a dedicated experiment. We start by defining reference frames in a sequence by applying the same criteria to create new reference frames as shown in the full pipeline (cf. Fig.~\ref{fig: flowchart}). The residuals are calculated using the ground truth relative pose between the reference frame and the current frame. The residuals are collected over an entire sequence, and then summarized in a histogram as shown in Fig.~\ref{fig: fitted sensor model}. By fitting the various distribution models depicted in Fig.~\ref{fig: Robust weight function} to the data, we finally identify the t-distribution to be the best model to describe the residual statistics. Assuming the mean of the t-distribution is always zero, only two parameters ($\nu_0$ and $\sigma_0$) have to be determined during the model fitting. As shown in~\cite{kerl2013robust}, the variance $\sigma$ is later on recursively updated on the actual data before being used for calculating weights.
% %
% \begin{equation}
% \sigma_{j+1}^{2} = \frac{1}{n}\sum_{i}r_{i}^2\frac{\nu + 1}{\nu + (\frac{r_i}{\sigma_j})^{2}},
% \end{equation}
% %
% where $j$ denotes the number of the iteration. However, in our experiment we find $\sigma$ should always be reset to the initial value $\sigma_0$ at the beginning of each iteration and the updated once value $\sigma_1$ is used for calculating the weight. The reason is that we observe the distribution of the residual in each iteration. It changes so much that the updated $\sigma$ in the last iteration is no longer a proper initial value for the current iteration.

%%%%%%%%%%%%%%%%%%%%%%%%%%%%%%%%%%%%%%%%%%%%%%%%%%%%%%%%%%%%%%%%%%%%%%%%%%%%%%%%
\section{Framework overview}
\label{Sec: Framework overview}

Here we discuss how to improve the robustness of the method further by incorporating a constant velocity motion model. Finally, we describe the complete VO pipeline.
%and provide an flowchart in Fig.~\ref{fig: flowchart}.

%++++++++++++++++++++++++++++++++++++++++++++++
\subsection{Constant velocity motion model}
\label{Subsec: Constant velocity motion model}

Given a sufficiently high processing rate, even a simple motion model can be very helpful to predict a good starting point for the optimization which is relatively close to the optimum where the residuals are minimal. This strategy has been widely used in VO and SLAM work~\cite{klein2007parallel,tanskanen2013live,kerl2013robust}, and it improved the robustness of the system by effectively avoiding local minima in the optimization. Instead of assuming a prior distribution for the motion as in~\cite{kerl2013robust}, we follow~\cite{klein2007parallel} and implement a simple decaying velocity model. It effectively improves the convergence speed and the tracking robustness, especially when the displacement between the reference frame and the current frame is relatively large.

%\begin{table*}[h]
%\centering
%\vspace{2mm}
%\begin{tabular}{|c|c|c|c|}
%	\hline
%    Robust Function & $h(r)$ & $w(r)$ & Parameter\\
%    \hline
%	$L_1$ & $\vert r \vert $ & $\frac{1}{\vert r \vert}$ & -- \\ \hline
%	$L_2$ & $\frac{r^2}{2}$ & $1$ & -- \\ \hline
%	Tukey & $
%    \left\{
%    \begin{array}{ll}
%    \frac{b^2}{6}(1-( 1-\frac{r^2}{b^2} ) ^3), \textup{if} \vert r\vert \leq b\\
%    \frac{b^2}{6}, \textup{if} \vert r \vert > b
%    \end{array}
%    \right.
%  $ & $\left\{
%    \begin{array}{ll}
%    (1-\frac{r^2}{b^2})^{2}, \textup{if} \vert r\vert \leq b\\
%    0, \textup{if} \vert r \vert > b
%    \end{array}
%    \right.$ & b = 4.6851\\ \hline
%    Cauchy & $\frac{b^2}{2}\log (1+\frac{r^2}{b^2})$ & $\frac{1}{1+r^2/b^2}$ & b = 2.3849\\ \hline
%    Huber & $\left\{
%    \begin{array}{ll}
%    \frac{r^2}{2}, \textup{if} \vert r\vert \leq b\\
%    b \vert r \vert -\frac{b^2}{2}, \textup{if} \vert r \vert > b
%    \end{array}
%    \right.$ & $\left\{
%    \begin{array}{ll}
%    1, \textup{if} \vert r\vert \leq b\\
%    \frac{b}{\vert r \vert}, \textup{if} \vert r \vert > b
%    \end{array}
%    \right.$ & b = 1.345\\ \hline
%    t-distribution & $\frac{((\nu - 1)/2)!}{2((\nu - 2)/2)!\,\sqrt[]{\pi\nu\sigma^2}}(1+\frac{r^2}{\nu \sigma^2})^{-\frac{\nu+1}{2}}$ & $\frac{\nu + 1}{\nu + (\frac{r}{\sigma})^2}$ & $\nu = 5, \sigma = \nu/(\nu - 2) $\\ \hline
%\end{tabular}
%\caption{Robust M-estimators and corresponding weighted functions.}
%\vspace{-5mm}
%\label{table: robust weight functions}
%\end{table*}

%+++++++++++++++++++++++++++++++++++++++++++++++++++++++++++++++++++++++++++++++
\subsection{Complete VO system}

\begin{figure}[H]
  \centering
  \includegraphics[width=0.8\columnwidth]{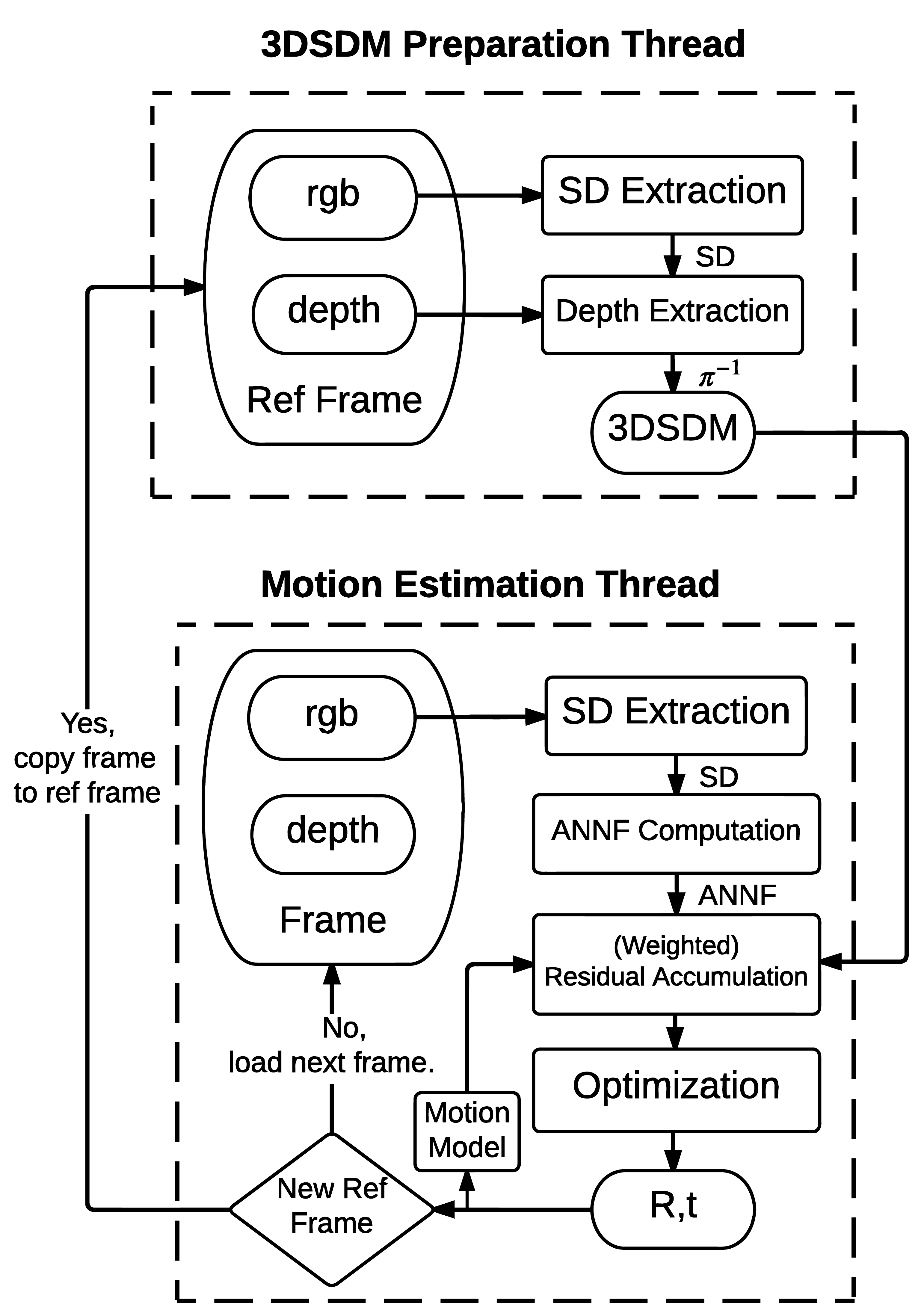}
  \caption{The flowchart of the proposed method. Two main threads are marked with dashed lines, in which processes are specified. SD refers to semi-dense region, 3DSDM represents 3D semi-dense map and ANNF means approximate nearest neighour field.}
%   \vspace{-5mm}
  \label{fig: flowchart}
\end{figure}

The complete VO system is designed based on the above robust motion estimation method. Two main threads are running in parallel, which are marked with dashed lines in Fig~\ref{fig: flowchart}. In the motion estimation thread, only the RGB image is used for the extraction of the semi-dense region and the subsequent ANNF computation. The objective is constructed and then optimized via the Gauss-Newton method. The reference frame needs to be updated once the current frame is too far away. Thus, we track the disparity between the semi-dense region in the reference frame and the corresponding pixels in the registered current frame. If the median disparity is larger than a given threshold, a new reference frame is created by the 3D semi-dense map (3DSDM) preparation thread, in which the depth information is loaded and corrected by the foreground reasoning operation described in Section \ref{Subsec: Problem formulation}.%rectified for errors at depth discontinuities.

%%%%figure%%%%
%\vspace{-4mm}
%%%%%%%%%%%%%%%%%%%%%%%%%%%%%%%%%%%%%%%%%%%%%%%%%%%%%%%%%%%%%%%%%%%%%%%%%%%%%%%%
\section{Experimental Evaluation}
\label{Sec: Evaluation}

We now proceed to the evaluation of the proposed method. We start by exploring various configurations in which performance under different semi-dense region extractors and different weight functions are assessed and compared. Then we provide a comparison between standard Euclidean distance field based method and our method, showing the advantage of the ANNF. We furthermore evaluate our algorithm on a set of benchmark datasets and compare the performance with several state-of-the-art camera tracking solutions. Finally, a semi-dense reconstruction result of two indoor scenes is provided which qualitatively demonstrates that the proposed method is able to work in relatively large-scale environments.

Note that errors listed in all tables are given in terms of either root-mean-square error or median error. The unit for rotation and translation error are $\deg/s$ and $m/s$ respectively. The best performance is always highlighted in bold.

%+++++++++++++++++++++++++++++++++++++++++++++++++
\subsection{Performance: Different gradient extractors}
\label{sec: Performance: Different gradient extractors}
A good extraction of semi-dense region is key to good motion estimation accuracy. Thus, we provide a comparison in Table~\ref{table:Performance comparison table} where several different methods are applied for calculating the image gradients. \quotes{Smoothed} refers to a $5 \times 5$ Gaussian kernel, which is used for smoothing the image. \quotes{Gradient} refers to a Sobel-like gradient computation method which uses a $5 \times 5$ kernel. It shows the impact that each method makes on the accuracy v.s. required computational time. Note that the t-distribution based IRLS is used. Run-time for computing the semi-dense region is expressed in seconds.

% We find that the failure of tracking is always due to bad quality of semi-dense regions~\ref{fig:}. A global shutter camera with relatively high definition can help to avoid this problem.
%+++++++++++++++++++++++++++++++++++++++++++++++++
% \vspace{-0.3mm}
\subsection{Performance: Different weight functions}
\label{Performance: Different weight functions}
In order to confirm experimentally that the chosen weight function is optimal, we compare the performance for all robust weight functions over several sequences of the TUM benchmark datasets. Comparison on fr2/desk is provided as an example in Table~\ref{table:Robust weight functions comparison}. \quotes{Smoothed + Gradient} is used for extracting the semi-dense region. The run-time is counted in seconds and includes the extraction of the semi-dense region, the ANNF computation and the following optimization.

%%table: comparison
\begin{table}
  \centering
  \renewcommand{\arraystretch}{1.2}
  \begin{tabular}{|M{1.8cm}|c|c|c|}
    \hline
    \multirow{2}*{Method} & \multicolumn{3}{c|}{\text{fr2/desk}} \\ %& \multicolumn{3}{c|}{\textbf{fr2**}} & \multicolumn{3}{c|}{\textbf{fr3**}} \\
    % \hline
    % \textbf{Inactive Modes} & \textbf{Description}\\
    \cline{2-4}
    & RMSE$(\mathbf{R})$ & 
    RMSE$(\mathbf{t})$ &
    Run-time \\
%     \textbf{$\mathbf{rms(e_R)}$} & 
%     \textbf{$\mathbf{rms(e_t)}$} &
%     \textbf{$\mathbf{\emptyset Runtime}$} &
%     \textbf{$\mathbf{rms(e_R)}$} & 
%     \textbf{$\mathbf{rms(e_t)}$} &
%     \textbf{$\mathbf{\emptyset Runtime}$} \\
    %\hhline{~--}
    \hline
 	Sobel & 0.562 & 0.018 & \textbf{0.00824} \\ \hline % &  &  &  &  &  &  \\ \hline
    Smoothed + Sobel & 0.581 & 0.018 & 0.00948 \\ \hline % &  &  &  &  &  &  \\ \hline
    Gradient & 0.565 & 0.017 & 0.01720\\ \hline % &  &  &  &  &  &  \\ \hline
    Smoothed + Gradient & \textbf{0.560} & \textbf{0.016} & 0.01863 \\ \hline % &  &  &  &  &  &  \\ \hline
    
  \end{tabular}
  \caption{Different methods of semi-dense region extraction.}
  \label{table:Performance comparison table}
  
  \begin{tabular}{|M{1.8cm}|c|c|c|}
    \hline
    \multirow{2}*{Method} & \multicolumn{3}{c|}{\text{fr2/desk}} \\ % & \multicolumn{3}{c|}{\textbf{fr2**}} & \multicolumn{3}{c|}{\textbf{fr3**}} \\
    % \hline
    % \textbf{Inactive Modes} & \textbf{Description}\\
    \cline{2-4}
    & RMSE$(\mathbf{R})$ & 
    RMSE$(\mathbf{t})$ &
    Run-time \\
%     \textbf{$\mathbf{rms(e_R)}$} & 
%     \textbf{$\mathbf{rms(e_t)}$} &
%     \textbf{$\mathbf{\emptyset Runtime}$} &
%     \textbf{$\mathbf{rms(e_R)}$} & 
%     \textbf{$\mathbf{rms(e_t)}$} &
%     \textbf{$\mathbf{\emptyset Runtime}$} \\
    %\hhline{~--}
    \hline
	Least Squares & 0.899 & 0.024 & \textbf{0.03318} \\ \hline% &  &  &  &  &  &  \\ \hline
    $\ell_1$ norm& 0.587 & \textbf{0.016} & 0.04211 \\ \hline % &  &  &  &  &  &  \\ \hline
    Tukey & 0.879 & 0.023 & 0.04077\\ \hline % &  &  &  &  &  &  \\ \hline
	Huber & 0.591 & \textbf{0.016} & 0.04084\\ \hline % &  &  &  &  &  &  \\ \hline
    Cauchy & 0.769 & 0.019 & 0.04193\\ \hline % &  &  &  &  &  &  \\ \hline
	t-distribution & \textbf{0.560} & \textbf{0.016} & 0.04260\\ \hline % &  &  &  &  &  &  \\ \hline
    
  \end{tabular}
  \caption{Different robust weight functions.}
  \vspace{-10mm}
  \label{table:Robust weight functions comparison}
\end{table}
%+++++++++++++++++++++++++++++++++++++++++++++++++
\vspace{-0.3mm}
\subsection{Euclidean distance field  v.s. ANNF}

\begin{figure}[t]
  \centering
  \subfigure[The projection of the 3DSDM.]{
  \includegraphics[width=0.45\columnwidth]{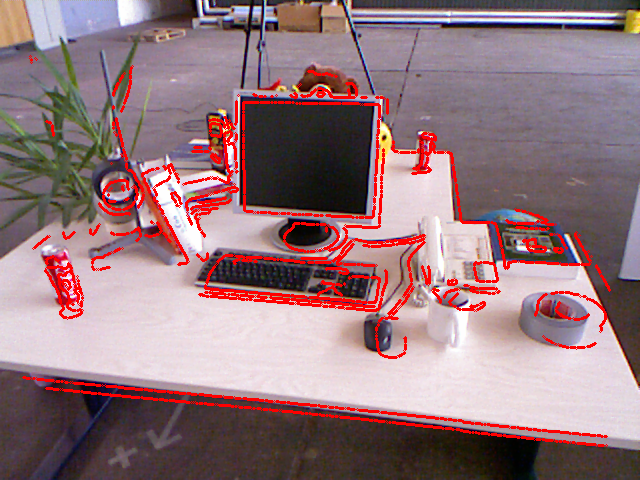}}
  \subfigure[Convergence speed.]{
  \includegraphics[width=0.45\columnwidth]{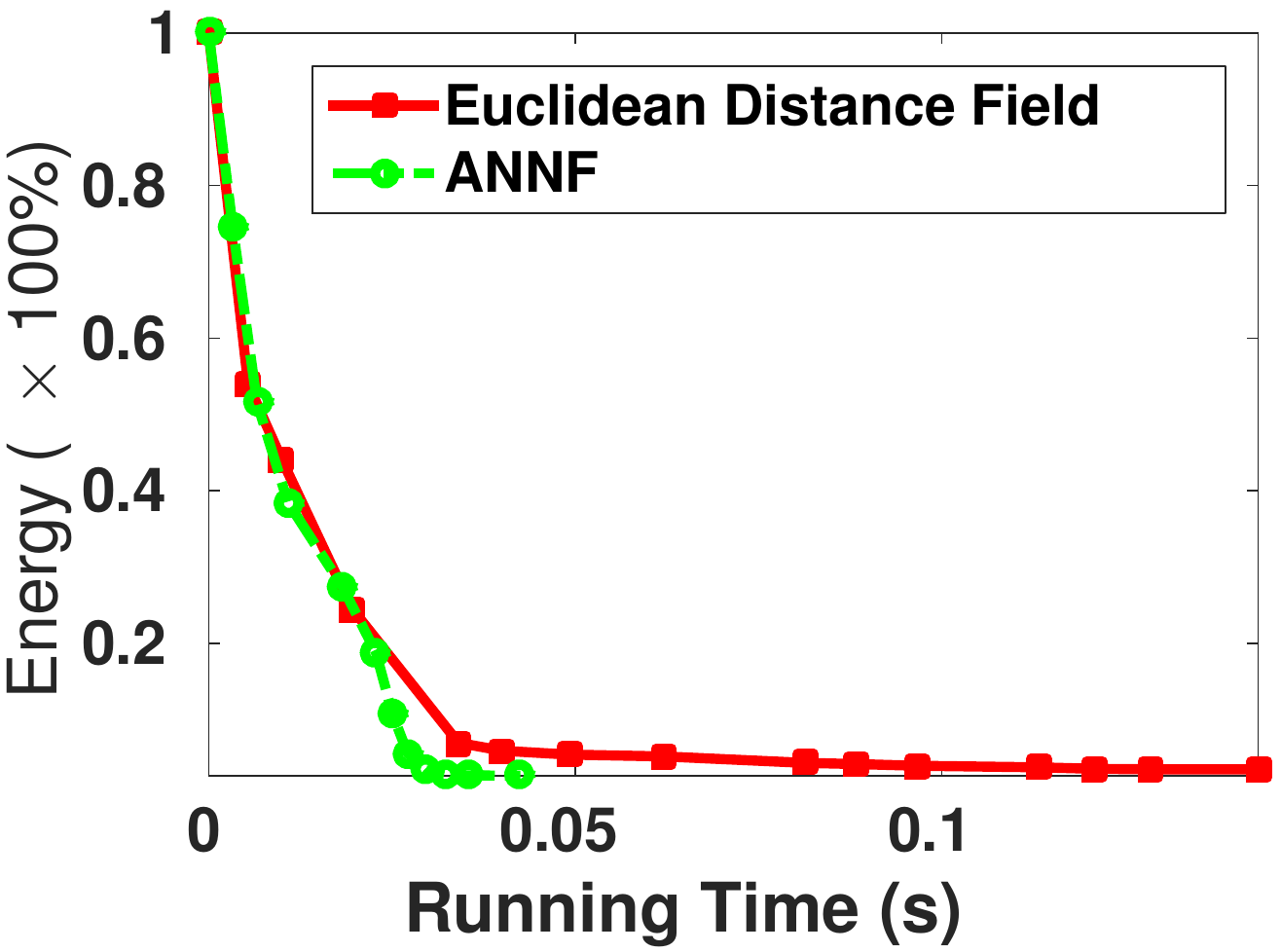}}
  \caption{Illustration of comparison between Euclidean distance based method and the novel ANNF based method. (a) shows the projection of the 3DSDM on the current frame before the optimization and (b) demonstrates the convergence speed of the two methods. Although both of them converge at similar rates, the ANNF based method is more efficient. The optimization of the two methods start from exact the same energy and normalized energy is depicted in (b).}
  \vspace{-5mm}
  \label{fig: convergence speed comparison}
\end{figure}

As discussed in~\ref{subsec: icp based motion estimation}, by using ANNF, we are able to calculate the signed orthogonal residual between the registered curves, thus enabling Gauss-Newton updates to solve the problem. Here we confirm that this leads to faster convergence compared to gradient descent (over the Euclidean distance field). The result in Fig.~\ref{fig: convergence speed comparison} demonstrates that ANNF based method converges much faster than standard Euclidean distance field based method. Note that \quotes{Smoothed + Gradient} and t-distribution based IRLS are used.
%%%%figure: sensor models%%%%
%+++++++++++++++++++++++++++++++++++++++++++++++++
\subsection{Comparison against the state of the art}

We compare the performance of our method against three state-of-the-art, open-source motion estimation frameworks: DVO~\cite{kerl2013robust}, LSD~\cite{engel2013semi,engel2014lsd} and ICP~\cite{whelan2012kintinuous}. %Both of them are direct methods. DVO is developed for RGB-D cameras and LSD is for monocular cameras. 
For best performance, we apply \quotes{Smoothed + Gradient} for extracting the semi-dense region, t-distribution based IRLS and enabled motion model. All methods are evaluated on published and challenging indoor benchmark datasets from the TUM RGB-D~\cite{sturm2012benchmark} series. %Besides, a comparison between ICP and our method on ICL-NUIM living room sequences (with noise) is also given.
The datasets we picked for evaluation and corresponding results are listed in Table~\ref{table: Performance comparison on TUM}. %and Table~\ref{table: Performance comparison on ICL} respectively.
DVO, LSD-SLAM and our method perform comparably efficient on a laptop with only CPU resources (30 Hz, 30 Hz and 25Hz respectively) while ICP achieves 60 Hz on a GPU (NVIDIA Tesla K40). It can be easily observed that our method provides the best overall performance on TUM dataset.% and reasonably good result on ICL-NUIM dataset\footnote{Our method also gives better or at least comparable results compared to \cite{kuse2016robust} and \cite{whelan2013robust} on several sequences of TUM dataset.}.% We do not present them all here because of insufficient space}. %and even outperforms dense photometric registration.

During the evaluation on the TUM dataset, we discovered that almost all underperforming registration results for our method are related to motion blur in the image. The reason is that semi-dense region cannot be accurately extracted from blurry images, thus also harming the resulting 3D semi-dense map. Consequently, the motion estimation based on the inaccurate 3D semi-dense map will not be accurate. Deblurring techniques or adaptive thresholds could alleviate this problem, but a much more straightforward solution consists of simply discarding frames for which the semi-dense region reveals a sudden jump in cardinality.

\begin{table*}[t!]
  \centering
  \renewcommand{\arraystretch}{1.2}
  \vspace{2mm}
  \begin{tabular}{|p{1.5cm}|c|c|c|c|c|c|c|c|c|c|}
    \hline
    \multirow{2}*{Odometer} & \multirow{2}*{Error}& \multirow{2}*{fr1/xyz} & \multirow{2}*{fr1/floor} & \multirow{2}*{fr2/xyz} & \multirow{2}*{fr2/rpy} & \multirow{2}*{fr2/desk} & \multirow{2}*{fr3/cabinet} & fr3/long office & fr3/structure & \multirow{2}*{Average} \\ 
    & & & & & & & & hosuehold & texture near & \\ \hline
			
	\multirow{4}{5em}{DVO} & RMSE(R) & 1.831 & 2.221 & 0.493 & 0.685 & 1.023 & 4.912 & 0.840 & \textbf{0.938} & 1.618\\
						   & Median(R) & \textbf{1.259} & 0.705 & 0.407 & 0.538 & 0.830 & 4.457 & 0.635 & \textbf{0.740} & 1.200\\
					       & RMSE(t) & 0.040 & 0.074 & 0.013 & 0.018 & 0.026 & 0.145 & 0.024 & \textbf{0.018} & 0.045\\
					       & Median(t) & 0.030 & 0.016 & 0.011 & 0.011 & 0.021 & 0.130 & 0.018 & \textbf{0.015} & 0.032\\ \hline
					       
	\multirow{4}{5em}{LSD SLAM} & RMSE(R) & 3.973 & 5.071 & 0.463 & 5.208 & 3.482 & 12.114 & \textbf{0.631} & 10.446 & 5.174\\
						   & Median(R) & 2.946 & 3.352 & 0.314 & 4.398 & 0.854 & 10.615 & 0.483 & 3.457 & 3.302\\
					       & RMSE(t) & 0.053 & 0.121 & 0.009 & 0.015 & 0.102 & 0.272 & 0.026 & 0.178 & 0.097\\
					       & Median(t) & 0.042 & 0.089 & 0.005 & 0.011 & 0.058 & 0.270 & 0.018 & 0.067 & 0.070\\ \hline
					       
	\multirow{4}{5em}{ICP} & RMSE(R) & 1.812 & 5.252 & 1.307 & 2.760 & 4.393 & 6.640 & 5.733 & 6.019 & 4.240\\
						   & Median(R) & 1.346 & 2.181 & 0.893 & 1.951 & 3.071 & 6.027 & 3.960 & 1.929 & 2.670\\
					       & RMSE(t) & \textbf{0.031} & 0.209 & 0.027 & 0.053 & 0.108 & 0.171 & 0.131 & 0.132 & 0.108\\
					       & Median(t) & \textbf{0.024} & 0.069 & 0.021 & 0.041 & 0.078 & 0.153 & 0.099 & 0.056 & 0.068\\ \hline
					       
	\multirow{4}{5em}{Our Method} & RMSE(R) & \textbf{1.533} & \textbf{0.746} & \textbf{0.324} & \textbf{0.356} & \textbf{0.560} & \textbf{2.692} & 0.673 & 1.128 & \textbf{1.001}\\
						   		  & Median(R) & 1.373 & \textbf{0.587} & \textbf{0.251} & \textbf{0.283} & \textbf{0.425} & \textbf{1.451} & \textbf{0.411} & 0.789 & \textbf{0.696} \\
					              & RMSE(t) & 0.041 & \textbf{0.013} & \textbf{0.005} & \textbf{0.005} & \textbf{0.016} & \textbf{0.063} & \textbf{0.016} & 0.024 & \textbf{0.023}\\
					        	  & Median(t) & 0.028 & \textbf{0.012} & \textbf{0.004} & \textbf{0.004} & \textbf{0.012} & \textbf{0.034} & \textbf{0.010} & 0.017 & \textbf{0.015}\\ \hline
    
  \end{tabular}
  \caption{Performance comparison on TUM dataset.}
  \vspace{-5mm}
  \label{table: Performance comparison on TUM}
\end{table*}

\subsection{Semi-dense reconstruction}

\begin{figure}[h]
  \centering
  \subfigure[TAMU corridor.]{
  \includegraphics[width=0.99\columnwidth]{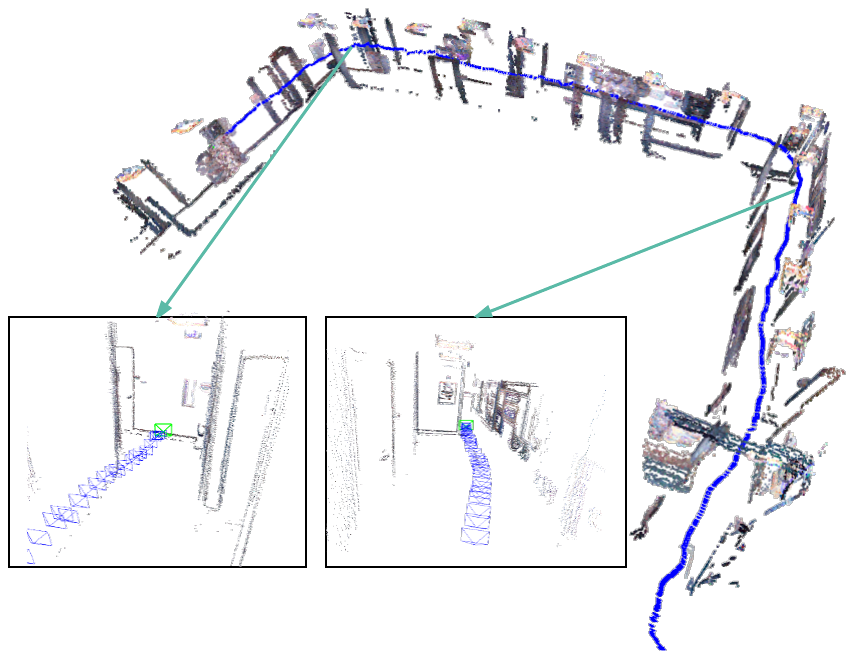}}
  \subfigure[ICL living room.]{
  \includegraphics[width=0.99\columnwidth]{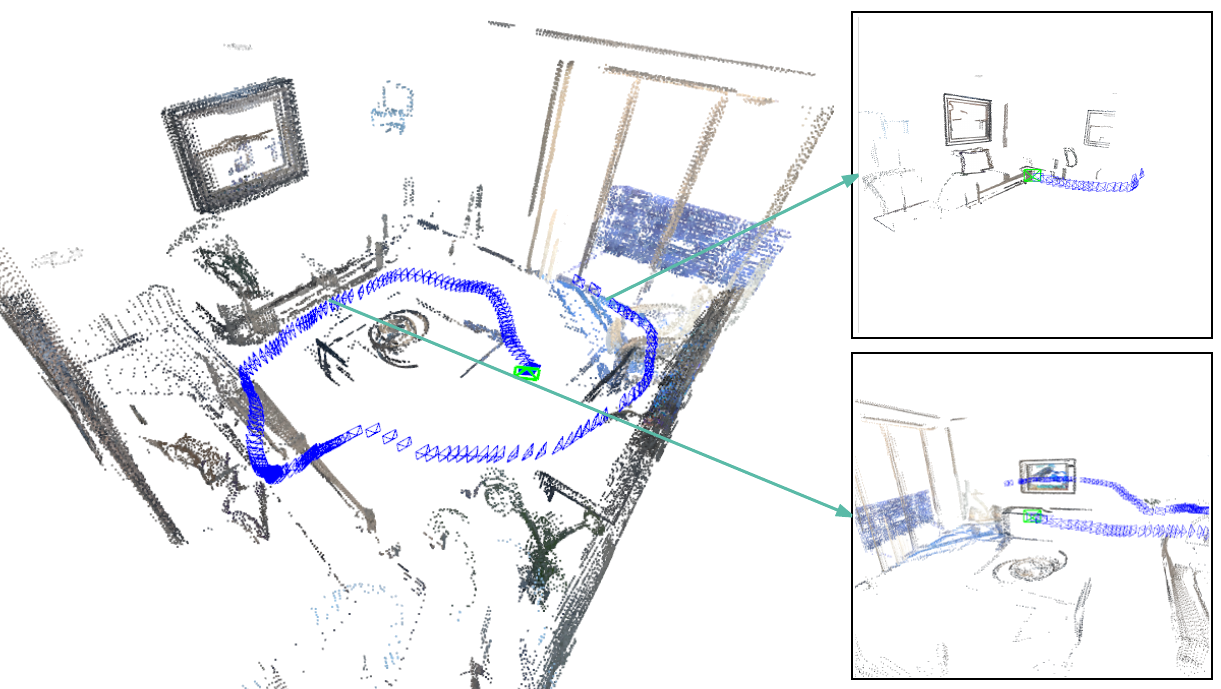}}
  \caption{Reconstruction result of indoor scenes.} %Close-up views show that the reconstruction of the semi-dense regions conveys a large amount of semantic information for high level manipulation. }
  \label{fig: semi dense reconstruction}
\end{figure}

In order to show that our method is capable to work in relatively large-scale environments, we provide reconstruction results on two sequences from the TAMU RGB-D datasets~\cite{lu2015robustness} and ICL-NUIM synthetic datasets~\cite{handa:etal:ICRA2014}. As shown in Fig.~\ref{fig: semi dense reconstruction}, the semi-dense reconstruction %conveys abundance of semantic information which could be used for high level applications such as manipulation and interaction.
is much more visually expressive than sparse point clouds. %It can furthermore be observed that our method has low drift, thus proving consistent reconstructions.
%
%%figure
%
%%%%%%%%%%%%%%%%%%%%%%%%%%%%%%%%%%%%%%%%%%%%%%%%%%%%%%%%%%%%%%%%%%%%%%%%%%%%%%%%

%%%%%%%%%%%%%%%%%%%%%%%%%%%%%%%%%%%%%%%%%%%%%%%%%%%%%%%%%%%%%%%%%%%%%%%%%%%%%%%%
%\vspace{-1mm}
\section{Conclusions}
We present a robust real-time semi-dense visual odometry algorithm for RGB-D cameras. The camera motion is estimated through a non-parametric 2D-3D geometric curve registration approach. The introduction of ANNF enabled the use of Gauss-Newton method. To improve robustness against occlusion, noises and outliers, the ICP-based pipeline is formulated as a maximum a posteriori problem, which is subsequently transformed into a weighted least squares problem. Furthermore, we explored a number of robust M-estimators by studying the statistical properties of the sensor model, and pick an adequate choice. Experiments show that our geometric registration alternative outperforms state-of-the-art camera tracking solutions in most cases. The method may be pushed even further by a more accurate and robust method for extracting contours, a more elaborate motion estimation filter, as well as a sliding windowed refinement. Hybrid cues (geometric and photometric) are viable for our framework and will be implemented in future work.
\bibliographystyle{IEEEtran}
\bibliography{IEEEabrv,myBib}

\begin{thebibliography}{10}
\providecommand{\url}[1]{#1}
\csname url@samestyle\endcsname
\providecommand{\newblock}{\relax}
\providecommand{\bibinfo}[2]{#2}
\providecommand{\BIBentrySTDinterwordspacing}{\spaceskip=0pt\relax}
\providecommand{\BIBentryALTinterwordstretchfactor}{4}
\providecommand{\BIBentryALTinterwordspacing}{\spaceskip=\fontdimen2\font plus
\BIBentryALTinterwordstretchfactor\fontdimen3\font minus
  \fontdimen4\font\relax}
\providecommand{\BIBforeignlanguage}[2]{{%
\expandafter\ifx\csname l@#1\endcsname\relax
\typeout{** WARNING: IEEEtran.bst: No hyphenation pattern has been}%
\typeout{** loaded for the language `#1'. Using the pattern for}%
\typeout{** the default language instead.}%
\else
\language=\csname l@#1\endcsname
\fi
#2}}
\providecommand{\BIBdecl}{\relax}
\BIBdecl

\bibitem{klein2007parallel}
G.~Klein and D.~Murray, ``Parallel tracking and mapping for small {AR}
  workspaces,'' in \emph{Mixed and Augmented Reality, 6th IEEE and ACM
  International Symposium on}, 2007, pp. 225--234.

\bibitem{mur2015orb}
R.~Mur-Artal, J.~Montiel, and J.~D. Tard{\'o}s, ``{ORB-SLAM}: a versatile and
  accurate monocular slam system,'' \emph{IEEE Transactions on Robotics},
  vol.~31, no.~5, pp. 1147--1163, 2015.

\bibitem{tykkala2011direct}
T.~Tykk{\"a}l{\"a}, C.~Audras, and A.~I. Comport, ``Direct iterative closest
  point for real-time visual odometry,'' in \emph{Computer Vision Workshops
  (ICCV Workshops), 2011 IEEE International Conference on}, pp. 2050--2056.

\bibitem{kerl2013robust}
C.~Kerl, J.~Sturm, and D.~Cremers, ``Robust odometry estimation for {RGB-D}
  cameras,'' in \emph{Robotics and Automation (ICRA), 2013 IEEE International
  Conference on}, pp. 3748--3754.

\bibitem{steinbrucker2011real}
F.~Steinbr{\"u}cker, J.~Sturm, and D.~Cremers, ``Real-time visual odometry from
  dense {RGB-D} images,'' in \emph{Computer Vision Workshops (ICCV Workshops),
  2011 IEEE International Conference on}, pp. 719--722.

\bibitem{audras2011real}
C.~Audras, A.~Comport, M.~Meilland, and P.~Rives, ``Real-time dense
  appearance-based {SLAM} for {RGB-D} sensors,'' in \emph{Australasian Conf. on
  Robotics and Automation}, vol.~2, 2011, pp. 2--2.

\bibitem{engel2013semi}
J.~Engel, J.~Sturm, and D.~Cremers, ``Semi-dense visual odometry for a
  monocular camera,'' in \emph{Proceedings of the IEEE International Conference
  on Computer Vision}, 2013, pp. 1449--1456.

\bibitem{engel2014lsd}
J.~Engel, T.~Sch{\"o}ps, and D.~Cremers, ``{LSD-SLAM}: Large-scale direct
  monocular {SLAM},'' in \emph{European Conference on Computer Vision}.\hskip
  1em plus 0.5em minus 0.4em\relax Springer, 2014, pp. 834--849.

\bibitem{newcombe2011kinectfusion}
R.~A. Newcombe, S.~Izadi, O.~Hilliges, D.~Molyneaux, D.~Kim, A.~J. Davison,
  P.~Kohi, J.~Shotton, S.~Hodges, and A.~Fitzgibbon, ``Kinectfusion: Real-time
  dense surface mapping and tracking,'' in \emph{Mixed and augmented reality
  (ISMAR), 10th IEEE international symposium on}, 2011, pp. 127--136.

\bibitem{whelan2012kintinuous}
T.~Whelan, M.~Kaess, M.~Fallon, H.~Johannsson, J.~Leonard, and J.~McDonald,
  ``Kintinuous: Spatially extended kinectfusion,'' in \emph{3rd RSS Workshop on
  RGB-D: Advanced Reasoning with Depth Cameras, (Sydney, Australia)}, 2012.

\bibitem{pomerleau2011tracking}
F.~Pomerleau, S.~Magnenat, F.~Colas, M.~Liu, and R.~Siegwart, ``Tracking a
  depth camera: Parameter exploration for fast {ICP},'' in \emph{2011 IEEE/RSJ
  International Conference on Intelligent Robots and Systems}, pp. 3824--3829.

\bibitem{pomerleau2013comparing}
F.~Pomerleau, F.~Colas, R.~Siegwart, and S.~Magnenat, ``Comparing {ICP}
  variants on real-world data sets,'' \emph{Autonomous Robots}, vol.~34, no.~3,
  pp. 133--148, 2013.

\bibitem{eade2009edge}
E.~Eade and T.~Drummond, ``Edge landmarks in monocular {SLAM},'' \emph{Image
  and Vision Computing}, vol.~27, no.~5, pp. 588--596, 2009.

\bibitem{lu2015robustness}
Y.~Lu and D.~Song, ``Robustness to lighting variations: An {RGB-D} indoor
  visual odometry using line segments,'' in \emph{Intelligent Robots and
  Systems (IROS), 2015 IEEE/RSJ International Conference on}, pp. 688--694.

\bibitem{klein2008improving}
G.~Klein and D.~Murray, ``Improving the agility of keyframe-based {SLAM},'' in
  \emph{European Conference on Computer Vision}.\hskip 1em plus 0.5em minus
  0.4em\relax Springer, 2008, pp. 802--815.

\bibitem{nurutdinova2015towards}
I.~Nurutdinova and A.~Fitzgibbon, ``Towards pointless structure from motion:
  {3D} reconstruction and camera parameters from general 3d curves,'' in
  \emph{Proceedings of the IEEE International Conference on Computer Vision},
  2015, pp. 2363--2371.

\bibitem{kuse2016robust}
M.~P. Kuse and S.~Shen, ``Robust camera motion estimation using direct edge
  alignment and sub-gradient method,'' in \emph{IEEE International Conference
  on Robotics and Automation (ICRA-2016), Stockholm, Sweden}, 2016.

\bibitem{yang2013goicp}
J.~Yang, H.~Li, and Y.~Jia, ``{Go-ICP}: Solving {3D} registration efficiently
  and globally optimally,'' in \emph{Proceedings of the 14th International
  Conference on Computer Vision (ICCV)}, 2013, pp. 1457--1464.

\bibitem{BMVC2015_100}
\BIBentryALTinterwordspacing
L.~Kneip, Z.~Yi, and H.~Li, ``{SDICP}: Semi-dense tracking based on iterative
  closest points,'' in \emph{Proceedings of the British Machine Vision
  Conference (BMVC)}, M.~W.~J. Xianghua~Xie and G.~K.~L. Tam, Eds.\hskip 1em
  plus 0.5em minus 0.4em\relax BMVA Press, September 2015, pp. 100.1--100.12.
  [Online]. Available: \url{https://dx.doi.org/10.5244/C.29.100}
\BIBentrySTDinterwordspacing

\bibitem{zhang1997parameter}
Z.~Zhang, ``Parameter estimation techniques: A tutorial with application to
  conic fitting,'' \emph{Image and Vision Computing}, vol.~15, no.~1, pp.
  59--76, 1997.

\bibitem{aftab2015convergence}
K.~Aftab and R.~Hartley, ``Convergence of iteratively re-weighted least squares
  to robust {M}-estimators,'' in \emph{2015 IEEE Winter Conference on
  Applications of Computer Vision}, pp. 480--487.

\bibitem{cayleyparameter}
A.~Cayley, ``About the algebraic structure of the orthogonal group and the
  other classical groups in a field of characteristic zero or a prime
  characteristic,'' in \emph{Reine Angewandte Mathematik}, 1846.

\bibitem{fabbri20082d}
R.~Fabbri, L.~D.~F. Costa, J.~C. Torelli, and O.~M. Bruno, ``{2D} euclidean
  distance transform algorithms: A comparative survey,'' \emph{ACM Computing
  Surveys (CSUR)}, vol.~40, no.~1, p.~2, 2008.

\bibitem{tarrio15}
J.~J. Tarrio and S.~Pedre, ``Realtime edge-based visual odometry for a
  monocular camera,'' in \emph{IEEE International Conference on Computer Vision
  (ICCV)}, 2015, pp. 702--710.

\bibitem{tanskanen2013live}
P.~Tanskanen, K.~Kolev, L.~Meier, F.~Camposeco, O.~Saurer, and M.~Pollefeys,
  ``Live metric {3D} reconstruction on mobile phones,'' in \emph{Proceedings of
  the IEEE International Conference on Computer Vision}, 2013, pp. 65--72.

\bibitem{sturm2012benchmark}
J.~Sturm, N.~Engelhard, F.~Endres, W.~Burgard, and D.~Cremers, ``A benchmark
  for the evaluation of {RGB-D} {SLAM} systems,'' in \emph{2012 IEEE/RSJ
  International Conference on Intelligent Robots and Systems}, pp. 573--580.

\bibitem{handa:etal:ICRA2014}
A.~Handa, T.~Whelan, J.~McDonald, and A.~Davison, ``A benchmark for {RGB-D}
  visual odometry, {3D} reconstruction and {SLAM},'' in \emph{IEEE Intl. Conf.
  on Robotics and Automation, ICRA}, Hong Kong, China, May 2014.

\end{thebibliography}
\end{document}